\documentclass[journal]{IEEEtran}
\usepackage{amsmath,amsfonts,amssymb}
\usepackage{algorithmic}
\usepackage{array}
\usepackage[caption=false,font=normalsize,labelfont=sf,textfont=sf]{subfig}
\usepackage{textcomp}
\usepackage{stfloats}
\usepackage{url}
\usepackage{verbatim}
\usepackage{graphicx, color}
\def\BibTeX{{\rm B\kern-.05em{\sc i\kern-.025em b}\kern-.08em
    T\kern-.1667em\lower.7ex\hbox{E}\kern-.125emX}}
\usepackage{balance}
\usepackage{lineno}
\usepackage[export]{adjustbox}
\usepackage{makecell}
\usepackage[table]{xcolor} % Needed for cell coloring
\usepackage{paralist}
\usepackage{multirow}
\usepackage[bookmarks=false,colorlinks=true,linkcolor=red]{hyperref}
\usepackage{xcolor}

\begin{document}
\title{An Uncertainty-Aware Generalization Framework for Cardiovascular Image Segmentation}
% \author{Anonymous Authors%
% \thanks{Manuscript submitted for review.}}
\author{
Ting Yu Tsai,
Liangqiao Gui,
Yineng Chen,
Li Lin,
Shu Hu,
Connie W. Tsao,
Xin Li,
Shao Lin, \\
Ming-Ching Chang$^*$,
Hongtu Zhu,
Xin Wang$^*$\thanks{$^*$Corresponding author. Email: {mchang2, xwang56}@albany.edu.}

\thanks{Ting Yu Tsai, Liangqiao Gui, Yineng Chen, Xin Li, Shao Lin, Ming-Ching Chang, and Xin Wang are with the University at Albany, State University of New York, Albany, NY 12222, USA. (Emails: \{ttsai2, lgui, ychen77, xli48, slin, mchang2, xwang56\}@albany.edu)}

\thanks{Li Lin and Shu Hu are with Purdue University, West Lafayette, IN 47907, USA. (Emails: \{lin1785, hu968\}@purdue.edu)}

\thanks{Connie W. Tsao is with Harvard Medical School, Boston, MA 02115, USA and with Beth Israel Deaconess Medical Center, Boston, MA 02215, USA. (Email: ctsao1@bidmc.harvard.edu)}

\thanks{Hongtu Zhu is with the University of North Carolina at Chapel Hill, Chapel Hill, NC 27599, USA. (Email: htzhu@email.unc.edu)}
}
%%%%%%%%%%%%%%%%%%%%%%%%%%

\maketitle

\begin{abstract}
Deep learning models have achieved significant success in segmenting cardiovascular structures, but there is a growing need to improve their generalization and robustness. 
Current methods often face challenges such as overfitting and limited accuracy, largely due to their reliance on large annotated datasets and limited optimization techniques.
This paper introduces the UU-Mamba model, an extension of the U-Mamba architecture, designed to address these challenges in both cardiac and vascular segmentation. 
By incorporating Sharpness-Aware Minimization (SAM), the model enhances generalization by seeking flatter minima in the loss landscape. 
Additionally, we propose an uncertainty-aware loss function that integrates region-based, distribution-based, and pixel-based components, improving segmentation accuracy by capturing both local and global features.
We expand our evaluations on the ImageCAS (coronary artery) and Aorta (aortic branches and zones) datasets, which present more complex segmentation challenges than the ACDC dataset (left and right ventricles) used in prior work, showcasing the model's adaptability and resilience.
Our results confirm UU-Mamba's superior performance compared to leading models such as TransUNet, Swin-Unet, nnUNet, and nnFormer. We also provide a more in-depth assessment of the model's robustness and segmentation accuracy through extensive experiments.
\end{abstract}


\section{Introduction}

% (1) background, problem
\IEEEPARstart{B}{iomedical} image segmentation is crucial for medical image analysis, as it enables the precise identification and delineation of anatomical structures and abnormalities~\cite{zengxin2024}.
Accurate segmentation of cardiovascular structures, such as the heart, aorta, and coronary arteries, from Magnetic Resonance Imaging (MRI) and Computed Tomography (CT) images is essential for diagnosing a wide range of cardiovascular conditions, developing treatment plans, and evaluating therapeutic outcomes~\cite{bernard2018deep, litjens2017survey}.
Although MRI and CT provide high-resolution images that reveal detailed information about the structure, function, and composition of these cardiovascular regions, manual segmentation of these structures is time-consuming, labor-intensive, and subject to variability among observers. This limitation highlights the need for automated segmentation techniques that can improve consistency and efficiency in the analysis~\cite{petitjean2011review, maier2019gentle}.

% (2) challenge
The variability in cardiovascular anatomy, pathological changes, and imaging artifacts poses significant challenges for accurate MRI and CT image segmentation~\cite{li2018}. 
Conventional techniques, such as thresholding and edge detection, often fail to accurately capture the complex morphology of the heart, aorta, and coronary arteries. 
% (3) existing methods drawbacks
Recent advances in machine learning, particularly through Convolutional Neural Networks (CNNs)~\cite{lecun1995} and other deep learning models, have shown promise in addressing these challenges by learning intricate patterns from large datasets~\cite{fahmy2019}. 
% drawbacks
However, despite these advances, current deep learning models still require substantial computational resources and large annotated datasets, while their ability to generalize across diverse patient populations and imaging conditions remains limited~\cite{litjens2017survey}.

%%=========================
\begin{figure}[t]
\centerline{\includegraphics[width=\linewidth]{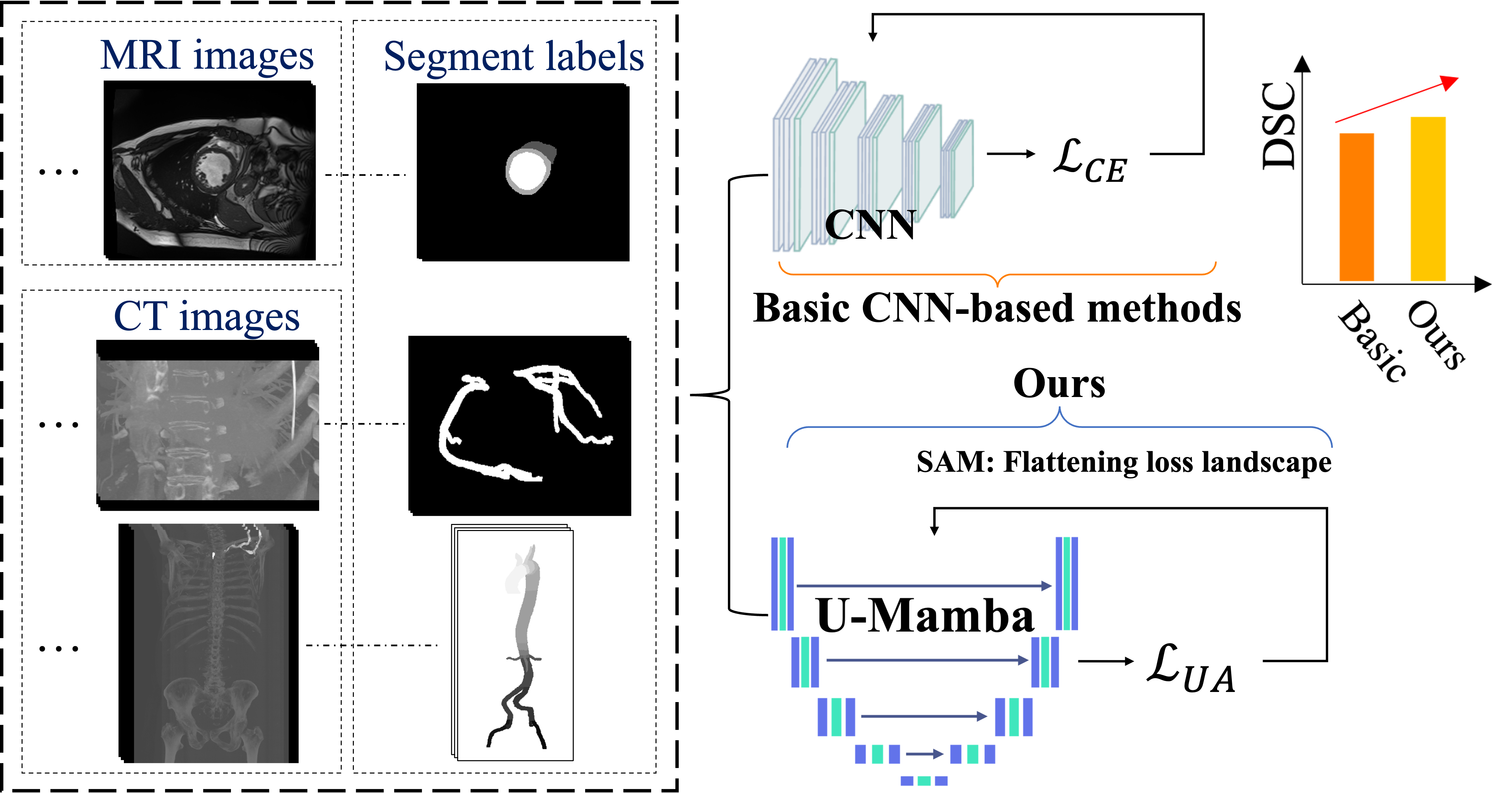}}
\caption{
Comparison between our method and basic approach. Traditionally, a deep learning model is trained using the Cross-Entropy loss $\mathcal{L}_{CE}$.
Our method enhances U-Mamba by utilizing the uncertainty-aware loss $\mathcal{L}_{UA}$, which is optimized via the SAM optimizer over a flattened loss landscape.
Evaluation using Dice Similarity Coefficient (DSC), Normalized Surface Dice (NSD) and Mean Squared Error (MSE) shows improvement of our method against basic CNN-based methods.}
\vspace{-4mm}
\label{fig:introduction}
\end{figure}
%%=========================

To improve segmentation accuracy and generalizability across various  cardiovascular structures, several specialized datasets have been established. 
For example, the Automated Cardiac Diagnosis Challenge (ACDC) dataset~\cite{bernard2018deep} focuses on cardiac segmentation, including the left and right ventricles and the myocardium in MRI images. The ImageCAS dataset~\cite{zeng2023imagecas} targets coronary artery segmentation, which is essential for the evaluation of coronary artery disease. The Aorta dataset~\cite{imran2024cis, krebs2024volumetric} specifically targets the segmentation of the aorta and its branches, supporting the diagnosis of conditions such as aortic aneurysms and dissections. These datasets collectively offer diverse challenges that contribute to the development of more robust and accurate segmentation algorithms.

% our methods,  summary tech
In this paper, we tackle these challenges by introducing an Uncertainty-Aware Generalization Framework, UU-Mamba, an enhanced version of the U-Mamba model~\cite{ma2024umamba}. 
The UU-Mamba model features a novel uncertainty-aware loss function along with the Sharpness-Aware Minimization (SAM) optimizer~\cite{foret2021sharpness}, which improves both training stability and performance. 
Our uncertainty-aware loss function is composed of three key components, inspired by a review of various loss functions for semantic segmentation tasks~\cite{jadon2020}. For example, \textit{Region-based loss}: Improves object detection and localization, such as Dice loss~\cite{sudre2017}. \textit{Distribution-based loss}: Compares predicted distributions with ground truth, typically using Cross-Entropy loss~\cite{ma2004}.
\textit{Pixel-based loss}: Measures differences at the pixel level, using techniques such as focal loss~\cite{lin2017}.
Instead of using fixed weights, our model dynamically adjusts the contribution of each loss component in accordance with the uncertainty of each prediction by employing auto-learnable weights~\cite{cipolla2018}. This enables the model to prioritize confident predictions while minimizing the influence of ambiguous or noisy data. The auto-learnable weights allow adaptive balancing of different aspects of the segmentation task, leading to improved overall performance and robustness, particularly in addressing challenges such as class imbalance~\cite{azad2023}. 
To further enhance our model's generalization capability, 
we implement the Sharpness-Aware Minimization (SAM) optimizer~\cite{foret2021sharpness}. SAM helps identify parameter values that lead to flatter minima in the loss landscape, improving the model's generalizability~\cite{lin2024robust,lin2024robust2,lin2024robust3,lin2024preserving} and reducing the risk of overfitting, a common challenge in deep learning applications for medical imaging~\cite{Mariam2024}.
Figure~\ref{fig:introduction} presents a comparison between our method and existing approaches.

%%%
% This paper extends our prior work~\cite{tsai2024uumamba} by comprehensively evaluating UU-Mamba model for both cardiac and vascular image segmentation tasks.
This paper extends prior work on UU-Mamba~\cite{tsai2024uumamba} by comprehensively evaluating the model for both cardiac and vascular image segmentation tasks. The main contributions are summarized as follows:

\begin{enumerate}% \begin{enumerate}[leftmargin=14pt] \itemsep -.1em
\item We expand the use of UU-Mamba model, originally developed for cardiac segmentation, to include vascular segmentation across multiple datasets. While the core architecture remains unchanged, we have enhanced its capabilities to address the unique challenges of both cardiac and vascular segmentation, demonstrating the model's versatility and improved performance across a broader range of medical imaging tasks.
% \item We present new comprehensive results on the ImageCAS~\cite{zeng2023imagecas} and Aorta~\cite{imran2024cis, krebs2024volumetric} datasets, which were not included in our prior work. These datasets differ in imaging characteristics and complexity. The Aorta dataset, for example, introduces a more intricate scenario with 24 labels.
% In comparison, the ACDC dataset~\cite{bernard2018deep} used in our prior work only involves three labels. This highlights the adaptability and robustness of our model across datasets with varying levels of complexity.
\item New results are reported on the ImageCAS~\cite{zeng2023imagecas} and Aorta~\cite{imran2024cis, krebs2024volumetric} datasets, which were not included in prior work. These datasets differ in imaging characteristics and label complexity. For example, the Aorta dataset includes 24 labels, whereas the ACDC dataset~\cite{bernard2018deep} used previously includes three labels, enabling assessment under substantially increased segmentation complexity.
% \item In addition to the Dice Similarity Coefficient (DSC) and Mean Squared Error (MSE) metrics used in our prior evaluation, we also incorporate Normalized Surface Dice (NSD)~\cite{maier-hein2022metrics}. We also include an analysis of the 3D loss landscape visualization~\cite{Hao2018}. These additional metrics provide a more comprehensive evaluation of the performance of the model, offering deeper insights into its ability to handle complex medical imaging tasks.
\item In addition to Dice Similarity Coefficient (DSC) and Mean Squared Error (MSE) used previously~\cite{tsai2024uumamba}, the evaluation incorporates Normalized Surface Dice (NSD)~\cite{maier-hein2022metrics} and analyzes the 3D loss landscape~\cite{Hao2018}. Together, these provide a more comprehensive characterization of model performance and optimization behavior on complex medical imaging tasks.
\end{enumerate}

%%%%%%%%%%%%%%%%%%%%%%%%%%%%%%%%%%%%%%
\section{Related Work}

%%%%%%%%%%%%%%%%%%%%%%%%%%%%%%%%%%%%%%%%%%%%%%%%%%%%%%%%%%%%%%%%%%%%%%%%%%%%
\subsection{Cardiovascular Image Segmentation}
\noindent The development of deep learning techniques, particularly Convolutional Neural Networks (CNNs)~\cite{lecun1995}, has significantly advanced cardiovascular segmentation \cite{wang2024artificial, zhou2019}, driven by comprehensive datasets such as ACDC~\cite{bernard2018deep}, ImageCAS~\cite{zeng2023imagecas}, and Aorta~\cite{imran2024cis, krebs2024volumetric}. These datasets address a range of cardiovascular structures, including heart chambers, coronary arteries, and aortic branches, and present challenges related to anatomical variability and disease-specific characteristics. Multi-modality cardiac imaging segmentation, which utilizes imaging modalities like Positron Emission Tomography (PET), Single Photon Emission Computed Tomography (SPECT), MRI, and CT, aims to precisely segment anatomical structures and pathological regions. However, inherent challenges such as phase alignment, resolution, and image quality imbalances complicate the process. Traditional methods, including registration-based segmentation with multi-atlas approaches, and fusion-based segmentation techniques, address these issues by combining information across modalities \cite{Zhuang_2019, zhao2022, eugenio2015, chen2020deep}, but these methods are computationally expensive and often require large datasets \cite{xu2023deep, Chartsias2021}. Hybrid approaches, which combine traditional and deep learning methods, are emerging to improve the robustness and clinical applicability of cardiovascular segmentation \cite{Fu_2020}.

Recent advances in deep learning have improved the precision of segmentation by enabling complex representations of cardiovascular structures. For example, CNNs such as U-Net~\cite{Isensee2021nnUNet, ronneberger2015u} and its variants have been effective for cardiovascular segmentation tasks, particularly on the ACDC dataset~\cite{bernard2018deep}. However, these methods are prone to overfitting and issues such as class imbalance, which complicates their application across diverse datasets like ImageCAS~\cite{zeng2023imagecas} and Aorta~\cite{imran2024cis, krebs2024volumetric}. Zeng \textit{et al.}~\cite{zeng2023imagecas} tackled coronary artery segmentation challenges in ImageCAS, using multi-scale feature extraction to capture finer details, but the method still struggles with arteries exhibiting atypical morphology, requiring precise hyperparameter tuning.

% The Aorta dataset~\cite{imran2024cis, krebs2024volumetric} focuses on segmenting the aorta and its branches, where the varying diameters of the aorta and pathologies such as aneurysms present additional complexities. Imran \textit{et al.}~\cite{imran2024cis} proposed CIS-UNet, a hybrid model incorporating Context-Aware Shifted Window Self-Attention mechanisms to address spatial variability in aortic structures, although its performance remains sensitive to image quality and resolution. To further improve segmentation, some studies have integrated CNNs with attention mechanisms and multi-scale processing. For example, Hu \textit{et al.}~\cite{hu2023} adapted the Segment Anything Model to medical images, incorporating multi-scale processing and CNN heads for enhanced segmentation across various datasets.

A key development in segmentation accuracy came from Isensee \textit{et al.}~\cite{Isensee2021nnUNet}, who combined U-Net and V-Net architectures in the nnUNet framework~\cite{Isensee2021nnUNet}. Although this method has proven effective, it is highly dependent on extensive annotated datasets, limiting its applicability in scenarios with scarce labeled data—a frequent issue in cardiovascular imaging. To address this, transfer learning has been employed, as demonstrated by Chen \textit{et al.}~\cite{chen2020deep}, who pre-trained models on the ACDC dataset and fine-tuned them on smaller, less-annotated datasets. Despite improving performance, transfer learning remains susceptible to domain shift issues, particularly when applied to datasets like ImageCAS and Aorta with differing data distributions.

Standard segmentation methods often rely on basic loss functions like Cross-Entropy loss, which struggle to effectively manage class imbalance or capture the finer details necessary for accurate segmentation. Recent research highlights the need to optimize for flatter minima in the loss landscape to enhance model generalization. Caldarola \textit{et al.}~\cite{caldarola2022} demonstrated that such optimization improves the robustness and generalization of the model, especially when dealing with noisy or ambiguous data, which is essential for reliable cardiovascular segmentation in diverse clinical scenarios.

\begin{figure*}[t]
    \centerline{
    \includegraphics[width=0.8\textwidth]{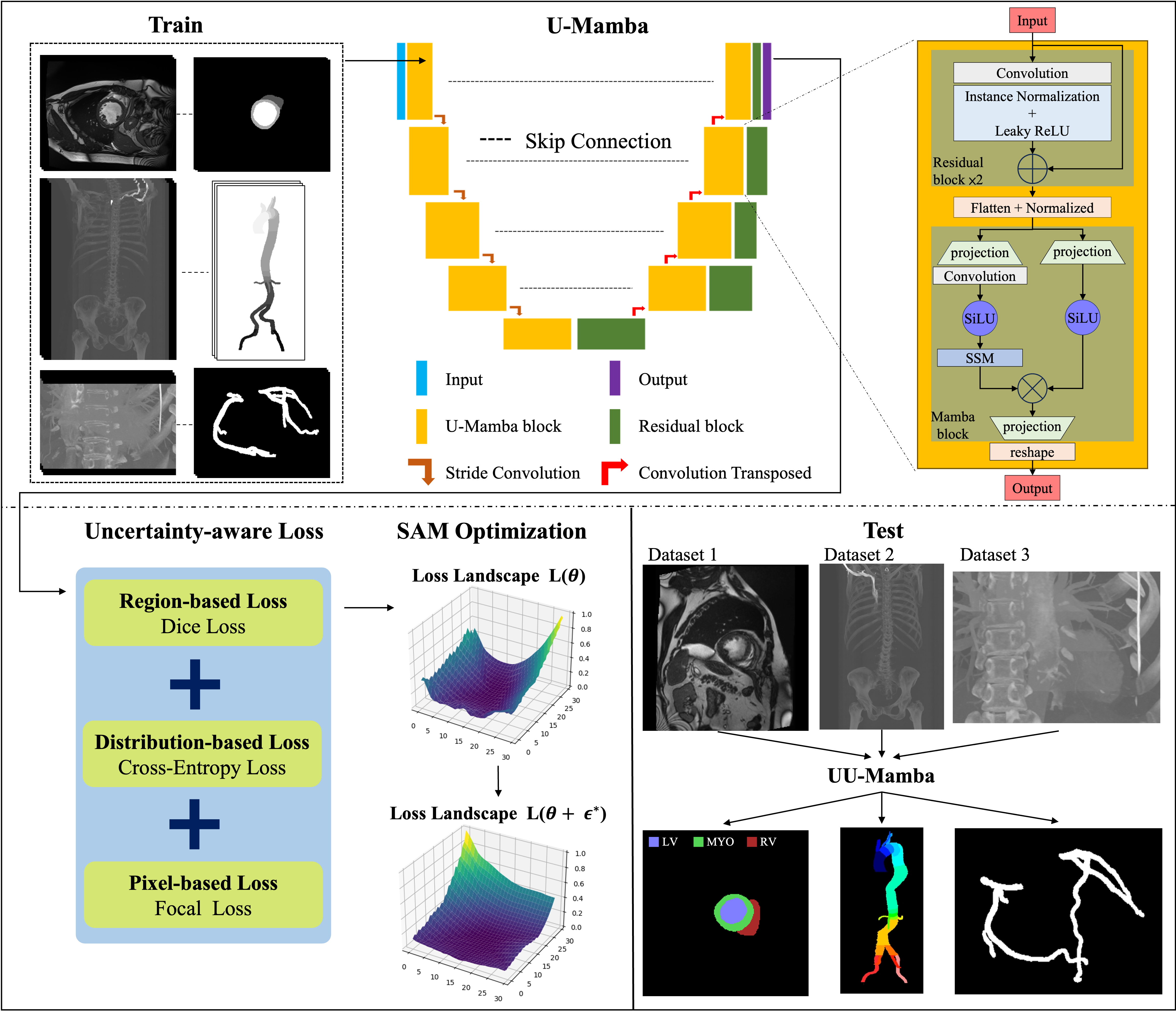}
    }
    \caption{Overview of our proposed framework, we encode input images and incorporate a novel uncertainty-aware loss function. 
    Optimization is performed using the Sharpness-Aware Minimization (SAM) optimizer~\cite{foret2021sharpness}, which operates within a flattened loss landscape. Experiments on the ACDC dataset~\cite{bernard2018deep}, ImageCAS dataset~\cite{zeng2023imagecas}, and Aorta dataset~\cite{imran2024cis, krebs2024volumetric} perform 3D heart segmentation on cardiovascular MRI and CT images, delineating each cardiovascular labels.
}
\vspace{-4mm}
\label{fig:Overview_proposed_model}
\end{figure*}
\subsection{Uncertainty-Aware Learning in Medical Image}

\noindent Understanding uncertainty is crucial for the medical image segmentation field, where the robustness and trustworthiness of predictions are essential in practical applications. The source of uncertainty can be mainly categorized into data uncertainty and model uncertainty. While data uncertainty is irreducible, model uncertainty can be reduced by improving the data ~\cite{gawlikowski2023survey, cipolla2018}.  Uncertainty-aware learning allows the model to adjust its training process based on the level of uncertainty, improving accuracy and robustness by emphasizing confident predictions and reducing the impact of uncertain ones ~\cite{cipolla2018}. Various techniques have been developed to incorporate uncertainty-aware learning.
Hu et al. ~\cite{hu2023umednerf} incorporate task-dependent uncertainty into MedNeRF to reconstruct 3D CT projections from a single X-ray view. By introducing uncertainty awareness, the loss terms of different tasks are adaptively weighted based on the levels of uncertainty of each task. Their experiment on chest and knee datasets shows that uncertainty-aware automatic weighting enhances both PSNR and SSIM compared to fixed-weight baselines, demonstrating that leveraging uncertainty-aware learning enables more stable convergence and improved reconstruction quality. 
Xia et al. ~\cite{xia2020uncertainty} develop an uncertainty-guided co-training (UMCT) framework that utilizes Bayesian dropout to estimate model uncertainty, which is then used to weight and fuse pseudo-labels, so that predictions with higher confidence contribute more to the training. This uncertainty-aware fusion improves the reliability of supervision on unlabeled data.
Nahiduzzaman et al. ~\cite{nahiduzzaman2025uncertainty} extend the Variational Information Pursuit (V-IP) framework by incorporating sample-specific uncertainty into concept-based reasoning. EUAV-IP skips unreliable concept predictions, while IUAV-IP prioritizes more reliable concepts in the query selection process, enabling the model to base its decisions on a smaller, more trustworthy subset of concepts while maintaining high diagnostic accuracy. Their models achieve improved predictive performance and shorter and meaningful explanations. 
Taken together, these studies indicate that uncertainty-aware learning is widely applied in medical image segmentation models and consistently outperforms conventional deterministic models under challenging conditions such as multi-task learning, limited labeled data availability, and noisy or ambiguous feature representations, by enabling models to adaptively focus on reliable information and reduce the impact of uncertain predictions.

\subsection{Sharpness-Aware Minimization in Medical Image Segmentation}

\noindent Recent studies have demonstrated that the geometry of the loss landscape plays a critical role in determining the generalization behavior of deep neural networks. Conventional optimization methods such as stochastic gradient descent (SGD) and Adam often converge to sharp minima that fit training data well but are highly sensitive to perturbations, resulting in limited robustness on unseen data. To address this issue, Sharpness-Aware Minimization (SAM) was introduced by Foret et al.~\cite{foret2021sharpness}, which explicitly seeks parameter configurations that minimize the worst-case loss within a local neighborhood of the solution. By encouraging convergence to flatter regions of the loss landscape, SAM improves model robustness and generalization. Recent theoretical analyses further confirm that SAM effectively reduces local sharpness and promotes flatter minima~\cite{wen2022does}.

SAM has attracted increasing attention in medical image analysis, where robustness and generalization are particularly important due to limited annotations, heterogeneous imaging protocols, and noisy labels. In retinal vessel segmentation, Mariam et al.~\cite{iqra2024} incorporated SAM into the RF-UNet framework and observed consistent improvements in accuracy, sensitivity, and specificity, alongside reduced training–validation performance gaps. In breast ultrasound segmentation, Hassan et al.~\cite{hassan2024} conducted a systematic evaluation of sharpness-based optimizers and reported that SAM consistently improved generalization across multiple architectures, outperforming several alternative sharpness-aware variants.

Beyond standard SAM, several extensions have been proposed to further enhance optimization stability and efficiency. Random Sharpness-Aware Minimization (RSAM) introduces stochastic perturbations to reduce computational overhead while maintaining robustness~\cite{yong2022rsam}, and Friendly SAM (FSAM) aims to improve optimization behavior in complex data regimes~\cite{tao2024friendly}. Although these variants have shown promising results in general vision tasks, their adoption in medical image segmentation remains limited.

Despite these advances, the role of sharpness-aware optimization has not been systematically investigated in cardiovascular image segmentation, which presents unique challenges including strong anatomical variability, thin and elongated structures (e.g., coronary arteries), complex branching patterns, and heterogeneous imaging modalities such as MRI and CT. These characteristics tend to produce highly irregular loss landscapes, making models particularly vulnerable to sharp minima and performance degradation under domain shift~\cite{caldarola2022}. This gap motivates further exploration of sharpness-aware optimization strategies tailored to cardiovascular segmentation scenarios.

\section{Method}

\subsection{Overview}

\noindent Figure~\ref{fig:introduction} presents a comparison of our method against basic approaches, highlighting the advancements made by our UU-Mamba model.
Figure~\ref{fig:Overview_proposed_model} illustrates our proposed UU-Mamba architecture, showcasing improvements in the training process. This model builds upon the foundational U-Mamba structure, where input images are effectively encoded. A key innovation in our approach is the integration of a novel uncertainty-aware loss function, designed to better capture and manage the inherent uncertainties in the segmentation task. To further enhance model performance, we employ the Sharpness-Aware Minimization (SAM) optimizer~\cite{foret2021sharpness}. This optimizer is particularly well-suited for our architecture as it operates within a flattened loss landscape, which helps in achieving more robust and generalized training outcomes. These enhancements make UU-Mamba a more effective and adaptable model for both cardiac and vascular segmentation tasks.
Section $\S$~\ref{sec:Mamba:UMamba} discusses the Mamba block and the U-Mamba network, with a focus on the integration of state space models and their effectiveness in capturing long-range dependencies.
In Section $\S$~\ref{sec:uncertainty-aware:loss}, we introduce our uncertainty-aware loss, detailing how it combines multiple loss functions to boost model performance and robustness.
Finally, Section$\S$~\ref{sec:SAM:opt} covers the Sharpness Aware Minimization Optimization, emphasizing its advantages in achieving flat minima in the loss landscape, thereby enhancing generalization and mitigating overfitting.

%%%%%%%%%%%%%%%%%%%%%%%%%%%%%%%%%%%%%%%%%%%%%%%%%%%%%%%%%%%%%%%%%%%%%%%%%%%%
\subsection{Network Architecture}
\label{sec:Mamba:UMamba}

\noindent The U-Mamba network is designed to improve the accuracy of medical image segmentation and improve global context comprehension by combining the assets of the Mamba block~\cite{gu2023mamba} and U-Net~\cite{Isensee2021nnUNet, ronneberger2015u}. The Mamba block, which is specifically engineered for Selective Structured State Space Sequence Models (S6), is particularly well-suited for medical imaging duties due to its exceptional ability to manage long-range dependencies and sequential information.

State Space Models (SSM)~\cite{gu2021combining} describe systems in terms of their internal states and observations over time, thereby facilitating effective sequence modeling through these underlying states. The fundamental form is denoted as follows: \(\mathbf{x}_t\) is the input state vector, \(\mathbf{u}_t\) is the control input, \(\mathbf{w}_t\) is the process noise, \(\mathbf{A}\) is the state transition matrix, and \(\mathbf{B}\) is the control input matrix.
\begin{equation}
    \begin{aligned}
        \mathbf{x}_{t+1} = \mathbf{A}\mathbf{x}_t + \mathbf{B}\mathbf{u}_t + \mathbf{w}_t.
    \end{aligned}
    \label{eq:SSMs}
\end{equation}
For observation \(\mathbf{y}_t\), calculated using the observation matrix \(\mathbf{C}\), feedthrough matrix \(\mathbf{D}\), and observation noise \(\mathbf{v}_t\), 
\begin{equation}
    \begin{aligned}
        \mathbf{y}_t = \mathbf{C}\mathbf{x}_t + \mathbf{D}\mathbf{u}_t + \mathbf{v}_t.
    \end{aligned}
    \label{eq:observation_at_timet}
\end{equation}
The S6 architecture advances traditional state space models by integrating selective attention mechanisms and structured parameterization. The selective attention mechanism can be represented as:
\begin{equation}
    \begin{aligned}
        \mathbf{a}_t = \text{softmax}(\mathbf{Q} \mathbf{K}^T / \sqrt{d_k}) \mathbf{V},
    \end{aligned}
    \label{eq:attention}
\end{equation}
where \(\mathbf{Q}\), \(\mathbf{K}\), and \(\mathbf{V}\) are the query, key, and value matrices that are derived from the state vector \(\mathbf{x}_t\), and \(d_k\) is the dimension of the key vectors. This mechanism enables the model to effectively capture intricate dependencies by concentrating on pertinent components of the input sequence.

The integration of S6 into the Mamba block is especially crucial for sequential medical image processing tasks, such as cardiac MRI segmentation, which require capture of temporal dynamics and structure~\cite{gu2023mamba}. The method, on the other hand, is exclusively concerned with per-image segmentation, which involves the application of the state transition and observation matrices (\(\mathbf{A}\), \(\mathbf{C}\), etc.) to individual images. Each image is treated independently.

U-Mamba capitalizes on Mamba's linear scaling advantage to improve CNNs' capacity to simulate long-range dependencies, all while circumventing the high computational costs associated with self-attention mechanisms employed in Transformers~\cite{vaswani2017attention} such as ViT~\cite{alexey2021} and SwinTransformer~\cite{liu2021swin}. The U-Mamba block, which is comprised of two sequential residual blocks followed by a Mamba block, is shown in Figure~\ref{fig:Overview_proposed_model}.

Additionally, each block includes Leaky ReLU activation, Instance Normalization, and convolutional layers. Mamba blocks with two parallel branches: one with an SSM layer and one without, flatten, transpose, normalize, and process image features. The Hadamard product is then employed to merge these features, which are subsequently projected back to their original shape and transposed.

An encoder with these blocks is included in the complete U-Mamba network architecture to capture both local features and long-range dependencies, while a decoder composed of residual blocks and transposed convolutions is used to recover detailed local information and resolution. Skip connections are used to connect hierarchical features from the encoder to the decoder. The final decoder output is processed through a 1$\times$1$\times$1 convolutional layer and a Softmax layer to generate the final segmentation probability map.

%%%%%%%%%%%%%%%%%%%%%%%%%%%%%%%%%%%%%%%%%%%%%%%%%%%%%%%%%%%%%%%%%%%%%%%%%%%%
\subsection{Loss Function}
\label{sec:uncertainty-aware:loss}

\noindent Introducing uncertainty into loss functions entails allocating weights to distinct components of the loss according to the estimated uncertainty for each data point~\cite{zhao2020uncertainty,zhao2019uncertainty}. 
This method allows the model to concentrate on learning from more dependable instances while simultaneously reducing the impact of potentially erroneous or ambiguous data. 
Kendall and Gal introduced the concept of adjusting loss functions by utilizing homoscedastic and heteroscedastic uncertainty~\cite{kendall2017uncertainties}. 
Heteroscedastic uncertainty fluctuates between instances, while homoscedastic uncertainty remains constant across all data points. 
The model can improve its resilience and precision by focusing on confident predictions and reducing the impact of equivocal ones by adapting its learning process to capitalize on these uncertainties. This optimization enhances overall performance and improves the training process across diverse datasets~\cite{hu2023umednerf,wang2024neural,peng2024uncertainty}.

To further boost segmentation accuracy, we employ an uncertainty-aware loss function that combines region-based, distribution-based, and pixel-based losses, capitalizing on their complementary strengths:
\begin{enumerate} %\begin{enumerate}[leftmargin=12pt] \itemsep -.1em
    \item {\bf Dice loss}~\cite{sudre2017}: This region-based metric emphasizes the overlap between predicted and ground truth areas, ensuring accurate preservation of shape and boundary details in segmented regions.
    \item {\bf Cross-Entropy (CE) loss}~\cite{ma2004}: This distribution-based loss ensures precise categorization of individual pixels, thereby improving classification accuracy.
    \item {\bf Focal loss}~\cite{lin2017}: This pixel-level loss addresses class imbalance by assigning greater importance to challenging instances, enhancing the model’s ability to manage complex scenarios~\cite{hu2020learning, hu2022sum, hu2023rank}.
\end{enumerate}

Let \(p_i\) denote the predicted probability and \(g_i\) the ground truth label, with the predicted segmentation and the corresponding ground truth mask. The Dice Similarity Coefficient (DSC) is a metric that quantifies the degree of overlap between the predicted segmentation and the ground truth. It is defined as follows:
\begin{equation}
    \begin{aligned}
        DSC = \frac{2 \sum_i p_i g_i}{\sum_i p_i + \sum_i g_i}
    \end{aligned}
    \label{eq:DSC}
\end{equation}

DSC values range from 0 to 1, with 1 signifying complete overlap between the prediction and the ground truth and 0 indicating no overlap.

The Dice loss is defined as: in order to integrate this metric into a loss function for training segmentation models.
\begin{equation}
    \begin{aligned}
        \mathcal{L}_{Dice} = 1 - DSC = 1 - \frac{2 \sum_i p_i g_i}{\sum_i p_i + \sum_i g_i}
    \end{aligned}
    \label{eq:DC_loss}
\end{equation}

The Dice loss is designed to minimize the discrepancy between the predicted segmentation and the ground truth by optimizing the DSC. The model is trained to generate segmentations that exhibit a greater overlap with the ground truth by minimizing the Dice loss, thereby enhancing the accuracy of the segmentation.

The standard entropy formula is employed to determine the Cross-Entropy (CE) loss:
\begin{equation}
    \begin{aligned}
        \mathcal{L}_{CE} = -\sum_i g_i \log(p_i)
    \end{aligned}
    \label{eq:CE_loss}
\end{equation}

To address class imbalance, we utilize the Focal loss, which focuses more on difficult-to-classify samples:
\begin{equation}
\begin{aligned}
    \mathcal{L}_{focal} = -\sum_i (1 - p_i)^\gamma g_i \log(p_i)
\end{aligned}
\label{eq:focal_loss}
\end{equation}
where \(\gamma\) is a focusing parameter default to 2.

The uncertainty-aware loss \(\mathcal{L}_{UA}\) is defined by combining these loss components within an uncertainty-aware framework:
\begin{equation}
    \begin{aligned}
        \mathcal{L}_{UA} = \sum_{m=1}^{M} \left( \frac{1}{2\sigma_m^2} \mathcal{L}_m + \log(1 + \sigma_m^2) \right)
    \end{aligned}
    \label{eq:uncertainty-aware_loss}
\end{equation}
in which \( M \) is the number of individual loss components, \( \mathcal{L}_m \) represents each loss component (such as Dice, CE, and Focal loss), and \( \sigma_m \) are learnable parameters that modify the contribution of each loss component based on the estimated uncertainty. To reduce the aggregate loss, these parameters are optimized during the training process.

By integrating uncertainty into the loss calculation, the model is able to dynamically modify the weights of each separate loss component. While mitigating the effects of class imbalance, this method strikes a balance between global and local accuracy. As an outcome, the model becomes more resilient to ambiguous or chaotic data, resulting in an overall improvement in segmentation performance.

%%%%%%%%%%%%%%%%%%%%%%%%%%%%%%%%%%%%%%%%%%%%%%%%%%%%%%%%%%%%%%%%%%%%%%%%%%%%
\subsection{Optimization Strategies for Model Generalization}
\label{sec:SAM:opt}

% Sharpness-Aware Minimization Optimization

\noindent To improve the U-Mamba model's generalizaiton in segmenting cardiovascular images, such as those in the ACDC~\cite{bernard2018deep}, ImageCAS~\cite{zeng2023imagecas}, and Aorta dataset~\cite{imran2024cis, krebs2024volumetric}, our methodology employs Sharpness-Aware Minimization (SAM) optimization~\cite{foret2021sharpness}. The model's generalizability is enhanced by the flattening of the loss landscape, implemented by SAM optimization. Despite the fact that conventional optimization techniques are designed to identify the lowest points in the loss landscape, these points are frequently precipitous, which results in inadequate generalization to new data. SAM, on the other hand, identifies gentler minima—regions in the parameter space where the model's performance remains consistent and is less susceptible to perturbations.

SAM is employed due to its effective reduction of overfitting, a common issue in medical image segmentation. Performance on unseen data may be impaired by the narrow valleys in the loss landscape that are a common consequence of conventional optimization methods. SAM, in contrast, concentrates on the identification of flattened minima, which are linked to enhanced generalization. When dealing with complex and diverse datasets, this is especially beneficial, as the variability in cardiac MRI images can exacerbate overfitting if not properly managed.

Optimization of SAM is accomplished through a two-step iterative procedure. Parameters are initially adjusted to optimize loss for each mini-batch. After this, the model parameters are adjusted to reduce the maximum loss. This perturbation is designed to identify model parameters that are located in flatter regions of the loss landscape, which are typically associated with improved generalization and increased robustness to minor changes in the input data.

The model parameters shall be denoted by \(\theta\), the loss function by \(\mathcal{L}\), and the training dataset by \(\mathcal{D}\). To investigate the loss landscape within a neighborhood around \(\theta\) defined by the norm constraint \(\|\epsilon\|_2 \leq \rho\), the perturbation \(\epsilon\) is introduced, with \(\rho\) determining the size of this neighborhood.

Mathematically, the SAM optimization is expressed as:
\begin{equation}
\begin{aligned}
    \theta^* = \arg \min_{\theta} \max_{\epsilon: \|\epsilon\|_2 \leq \rho} \mathcal{L}(\theta + \epsilon; \mathcal{D}).
\end{aligned}
\label{eq:SAM_optimization}
\end{equation}

The perturbation \(\epsilon\) within the \(\|\epsilon\|_2 \leq \rho\) constraint is determined in the initial phase to maximize the loss. This method identifies the worst-case direction in the local neighborhood of \(\theta\). Furthermore, this guarantees that the model parameters are directed toward regions of the loss landscape that are not precipitous. In order to enhance the parameters' resilience to perturbations, the model parameters \(\theta\) are modified in the second phase to reduce the loss at the worst-case perturbed location.

By applying these two stages iteratively, SAM directs the optimizer toward parameter configurations that are resilient to perturbations, thereby improving generalization and overall performance. The model is able to identify flattened minima in the loss landscape as a result of this approach, which results in improved generalization~\cite{foret2021sharpness}.

%-------------------------------------
\begin{figure*}[t]
    \centerline{\includegraphics[width=1\textwidth]{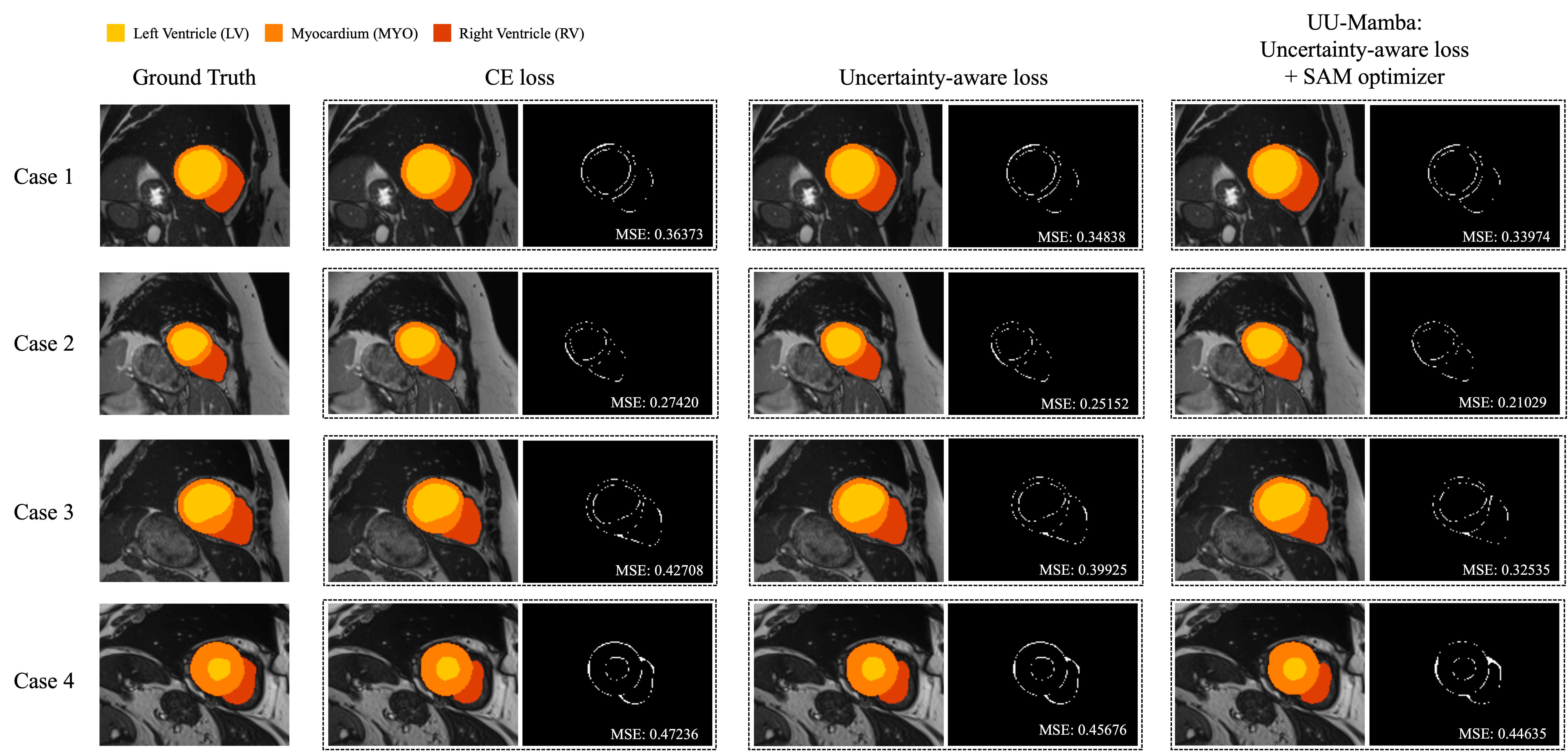}}  
    \caption{Segmentation results for various methods on sample images from the ACDC dataset~\cite{bernard2018deep}. The Mean Squared Error (MSE) between the output segmentation and the ground truth is shown for each method.}
    \vspace{-3mm}
    \label{fig:ACDC_result}
\end{figure*}
%-------------------------------------
%-------------------------------------
\begin{figure*}[t]
    \centerline{
    \includegraphics[width=0.8\textwidth]{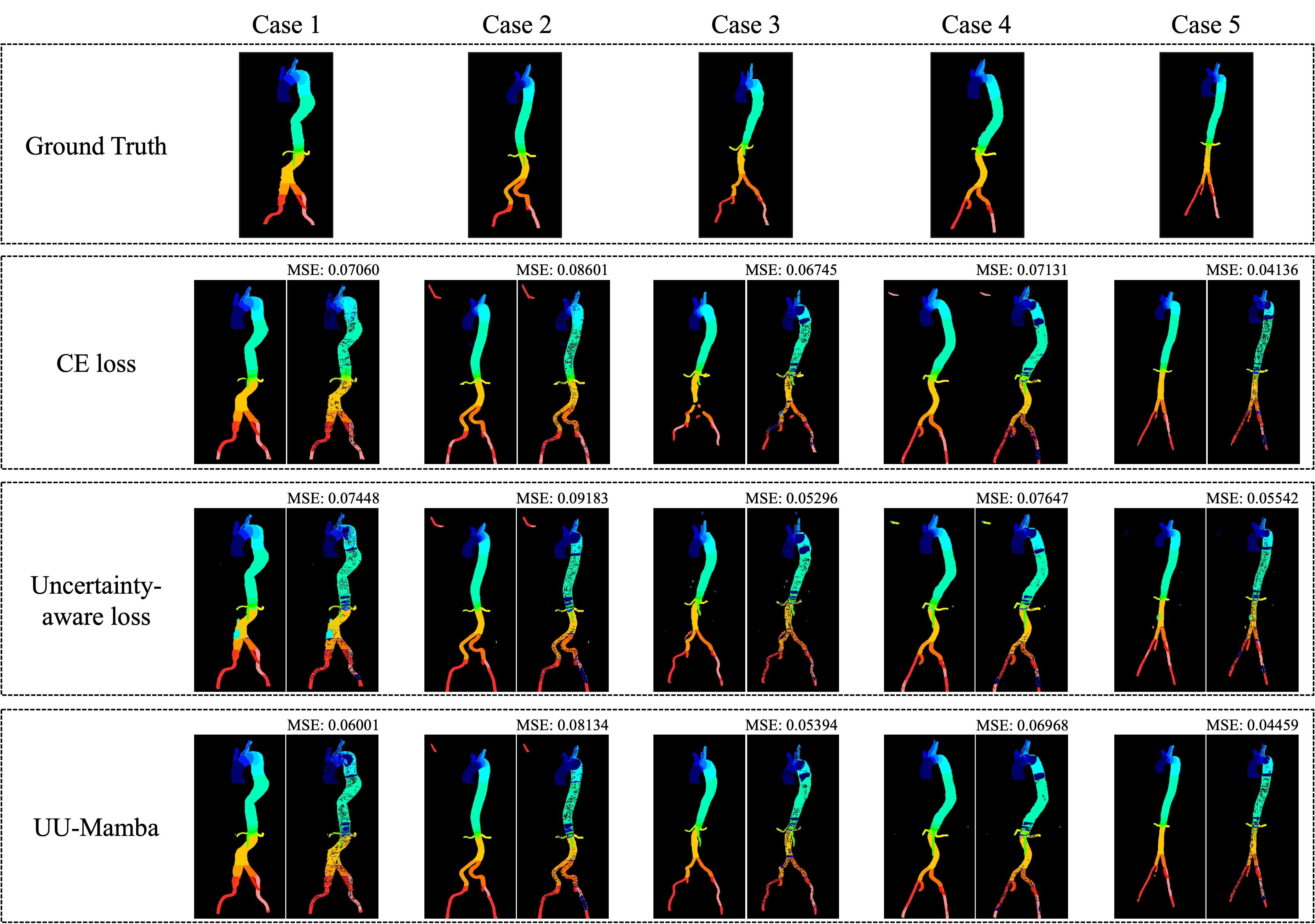}
    % \vspace{-2mm}
    }  
    \caption{Segmentation results for various methods on sample images from the Aorta dataset~\cite{imran2024cis, krebs2024volumetric}. The Mean Squared Error (MSE) between the output segmentation and the ground truth is shown for each method.}
    % \vspace{-3mm}
    \label{fig:Aorta_result}
\end{figure*}
%-------------------------------------
%-------------------------------------
\begin{figure*}[t]
    \centerline{
    \includegraphics[width=0.8\textwidth]{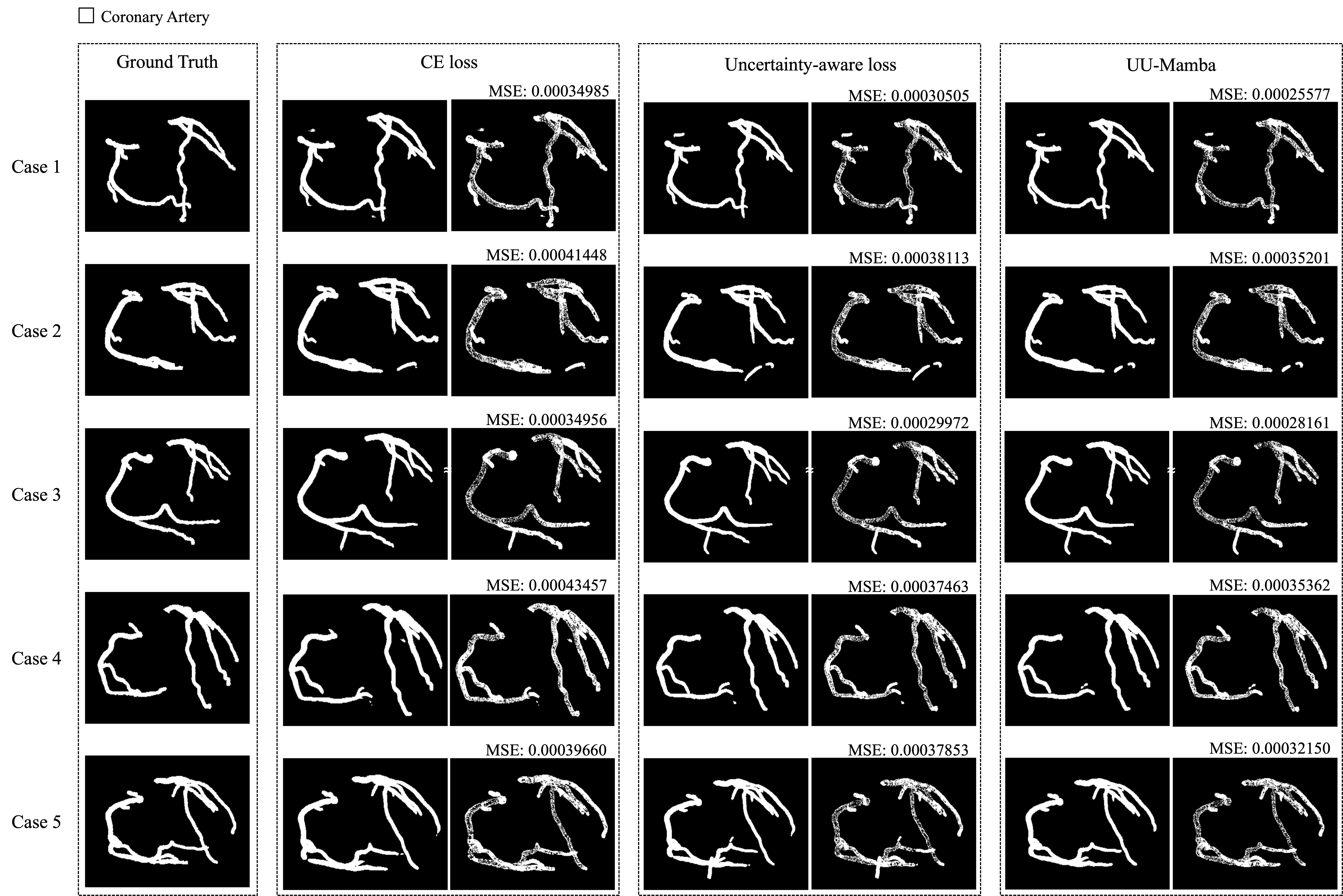}
    % \vspace{-2mm}
    }  
    \caption{Segmentation results for various methods on sample images from the ImageCAS dataset~\cite{zeng2023imagecas}. The Mean Squared Error (MSE) between the output segmentation and the ground truth is shown for each method.}
    \vspace{-3mm}
    \label{fig:ImageCAS_result}
\end{figure*}
%-------------------------------------

%%%%%%%%%%%%%%%%%%%%%%%%%%%%%%%%%%%%%%
\section{Experiments}

%%%%%%%%%%%%%%%%%%%%%%%%%%%%%%%%%%%%%%%%%%%%%%%%%%%%%%%%%%%%%%%%%%%%%%%%%%%%
\subsection{Datasets}

%%%%%%%%%%%%%%%%%%%%%%%%%%%%%%%%%%%%%%%%%%%%%%%%%%%%%%%%%%%%%%%%%%%%%%%%%%%%
\subsubsection{The ACDC Dataset}
\noindent The Automated Cardiac Diagnosis Challenge (ACDC) dataset~\cite{bernard2018deep} is a widely recognized benchmark in medical image analysis, particularly for cardiac MRI segmentation. With a total of 300 images and 2,978 slices, this dataset comprises MRI scans from 150 patients, each divided into numerous slices. {\em normal subjects}, {\em myocardial infarction}, {\em dilated cardiomyopathy}, {\em hypertrophic cardiomyopathy}, and {\em abnormal right ventricle} are the five distinct categories in which the patients are evenly distributed. Each group is distinguished by specific cardiac pathologies.

The dataset includes short-axis cardiac MRI images that provide a thorough examination of the heart. Ground truth annotations are supplied for the left ventricle (LV), right ventricle (RV), and myocardium (MYO) in each image, facilitating the formulation and assessment of segmentation algorithms. The ACDC dataset also demonstrates variability in both image spacing and size across various dimensions, which further complicates the segmentation task.

%%%%%%%%%%%%%%%%%%%%%%%%%%%%%%%%%%%%%%%%%%%%%%%%%%%%%%%%%%%%%%%%%%%%%%%%%%%%
\subsubsection{The Aorta Dataset}
\noindent The Aorta dataset~\cite{imran2024cis, krebs2024volumetric} is a meticulously annotated compilation of 50 Computed Tomography Angiography (CTA) images that enables the multi-class segmentation of the aorta and its branches. The axial dimensions of these images range from $389 \times 389$ pixels to $516 \times 516$ pixels, with an average size of $450 \times 450$ pixels. The dataset is guaranteed to be consistent in measurement, as each image maintains an isotropic voxel resolution of 1mm x 1mm x 1mm. The average number of axial segments per scan is 695, with image numbers ranging from 578 to 801. This dataset is essential for the development and testing of sophisticated algorithms that are designed to accurately segment the complex vascular structures within the aorta and its branches, thereby establishing a strong foundation for research in medical image analysis and machine learning.

%%%%%%%%%%%%%%%%%%%%%%%%%%%%%%%%%%%%%%
\subsubsection{The ImageCAS Dataset}
\noindent The ImageCAS dataset~\cite{zeng2023imagecas} focuses on the segmentation of coronary arteries using CTA images. This dataset contains approximately 1,000 3D CTA images, making it considerably larger than existing public datasets in this domain, which are crucial for diagnosing and assessing coronary artery disease. The dataset is particularly challenging due to the small size and complex branching patterns of the coronary arteries, as well as the motion artifacts introduced by cardiac and respiratory movements. Ground truth annotations include detailed segmentations of the coronary arteries, providing a comprehensive framework for evaluating the performance of segmentation algorithms in detecting and delineating these critical structures.

These datasets provide a diverse and comprehensive set of challenges for cardiac and vascular segmentation, allowing us to rigorously evaluate the effectiveness and generalizability of the proposed UU-Mamba model across different anatomical structures and imaging modalities.

\subsection{Experimental Settings}

%%%%%%%%%%%%%%%%%%%%%%%%%%%%%%%%%%%%%%
\subsubsection{Evaluation Metrics}
\noindent We employ the Dice Similarity Coefficient (DSC) as our primary metric to evaluate the performance of segmentation, according to the evaluation protocol described in \cite{maier-hein2022metrics}. The DSC assesses the overlap between the predicted segmentation and the ground truth mask, thereby providing a reliable indication of the model's accuracy in delineating cardiac structures. The DSC is defined in Eq.~\eqref{eq:DSC}.

Additionally, we employ the Mean Squared Error (MSE) to assess the average squared difference between the predicted probabilities and the ground truth labels, in addition to DSC. Let \(N\) denote the number of testing images, \(H\) and \(W\) denote the height and width of the images, \(p_{ij}^n\) be the predicted probability at pixel \((i, j)\) for the \(n\)-th image, and \(g_{ij}^n\) the corresponding ground truth label. The MSE is determined by the following formula:
\begin{equation}
\begin{aligned}
    \text{MSE} = \frac{1}{N} \sum_{n=1}^{N} \frac{1}{HW} \sum_{i=1}^{H} \sum_{j=1}^{W} (p_{ij}^n - g_{ij}^n)^2
\end{aligned}
\label{eq:MSE}
\end{equation}
MSE offers a complementary evaluation, providing insight into the model's pixel-wise precision.

We also incorporate the Normalized Surface Dice (NSD) metric that is recommended in~\cite{maier-hein2022metrics}. The NSD measures the average distance between the predicted segmentation surface and the ground truth surface, normalized by the ground truth surface area. This metric is particularly useful for assessing the spatial accuracy of the segmentation, especially in clinical scenarios where precise boundary delineation is critical. The NSD is defined as follows:
\begin{equation}
\begin{aligned}
    \text{NSD} = \frac{1}{|S_{gt}|} \sum_{x \in S_{gt}} \min_{y \in S_{pred}} \| x - y \|
\end{aligned}
\label{eq:NSD}
\end{equation}
where \(S_{gt}\) is the set of surface points in the ground truth segmentation, \(S_{pred}\) is the set of surface points in the predicted segmentation, and \(\| x - y \|\) represents the Euclidean distance between points \(x\) and \(y\). The NSD thus provides a detailed measure of the model's ability to accurately capture the shape and contours of the cardiac structures.

%%%%%%%%%%%%%%%%%%%%%%%%%%%%%%%%%%%%%%
\subsubsection{Implementation Details}
\noindent We conducted the experiments using the PyTorch framework and two NVIDIA A100 Tensor Core GPUs for training. During training, for the ACDC dataset~\cite{bernard2018deep}, we used a patch size of [20, 256, 224] and a batch size of 4, with the number of pooling operations per axis configured to [2, 5, 5]. In the case of the ImageCAS dataset~\cite{zeng2023imagecas}, a patch size of [96, 160, 160] and a batch size of 2 were selected, with pooling operations per axis set to [4, 5, 5]. For the Aorta dataset~\cite{imran2024cis, krebs2024volumetric}, the patch size was [176, 112, 112], also with a batch size of 2, and the pooling operations per axis were set to [4, 4, 4]. The network configuration comprises 6 stages. An initial learning rate of $5 \times 10^{-3}$ was utilized, and the training proceeded for 500 epochs. In the SAM optimization, the hyperparameter $\rho$ controlling the perturbation in Eq.~\eqref{eq:SAM_optimization} was set to 0.05. The focusing parameter $\gamma$ in the focal loss in Eq.~\eqref{eq:focal_loss} was set to 2. The parameter \( M \) of the uncertainty-aware loss in Eq.~\eqref{eq:uncertainty-aware_loss} was set to 3, incorporating Diclikee loss, Cross-Entropy loss, and Focal loss.

%-------------------------------------
\begin{table}[t]
\caption{
Performance comparison of our UU-Mamba with leading medical image segmentation methods on the ACDC dataset~\cite{bernard2018deep} for the three anatomical regions—the right ventricle (RV), left ventricle (LV), and myocardium (Myo). The evaluation metric is DSC (\%).}
\label{tab:Results_comparison_diff_models}
\vspace{-2mm}
\centerline{
%\normalsize
    \scalebox{1.0}{
    \begin{tabular}{c|c|c|c|c}
        \hline
        Method                             & Average        & RV $\uparrow$  & Myo $\uparrow$ & LV $\uparrow$  \\ \hline
        TransUNet \cite{Chen2021TransUNet} & 89.71          & 88.86          & 84.53          & 95.73          \\ \hline 
        Swin-Unet \cite{Cao2021SwinUnet}   & 90.00          & 88.55          & 85.62          & \textbf{95.83} \\ \hline
        nnUNet    \cite{Isensee2021nnUNet} & 91.61          & 90.24          & 89.24          & 95.36          \\ \hline
        nnFormer  \cite{Zhou2021nnFormer}  & 92.06          & 90.94          & 89.58          & 95.65          \\ \hline
        U-Mamba   \cite{ma2024umamba}      & 92.22          & 91.83          & 90.22          & 94.54          \\ \hline
        \textbf{Ours}           & \textbf{92.79} & \textbf{92.41} & \textbf{90.90} & 95.04          \\ \hline
    \end{tabular}
    }
}
\vspace{-1mm}
\end{table}
%-------------------------------------

%-------------------------------------
\begin{table}[t]
\caption{The U-Mamba backbone was employed to conduct an ablation study of ACDC dataset~\cite{bernard2018deep} for the three anatomical regions—the right ventricle (RV), left ventricle (LV), and myocardium (Myo). The study included the following configurations: (1) only the Cross-Entropy (CE) loss, (2) the uncertainty-aware loss without the SAM optimizer, and (3) the proposed UU-Mamba model (uncertainty-aware loss + SAM optimizer). DSC (\%) serves as the evaluation metric.}
\label{tab:Results_ablation_study_ACDC}
\vspace{-2mm}
\centerline{
    \renewcommand{\arraystretch}{1.3} % Increase row height
    % \normalsize
    \scalebox{0.8}{
    \begin{tabular}{c|c|c|c|c}
        \hline
        Method                                 & Avg. DSC \% $\uparrow$ & RV DSC \% $\uparrow$ & Myo DSC \% $\uparrow$ & LV DSC \% $\uparrow$ \\ \hline
        CE loss                                & 92.263                 & 91.81                & 90.31                 & 94.67                \\ \hline
        \makecell{Uncertainty-aware\\loss (CE, Dice, Focal)} & 92.602                 & 92.36                & 90.51                 & 94.94                \\ \hline
        \textbf{\makecell{Ours}} & \textbf{92.787}        & \textbf{92.41}       & \textbf{90.90}        & \textbf{95.04}       \\ \hline
    \end{tabular}
    }
}
\vspace{-1mm}
\end{table}
%-------------------------------------
%-------------------------------------
\begin{table}[t]
\caption{Ablation study of Aorta dataset~\cite{imran2024cis, krebs2024volumetric} on various configurations with U-Mamba backbone: (1) using only the Cross-Entropy (CE) loss, (2) using the uncertainty-aware Loss without the SAM optimizer, and (3) the proposed UU-Mamba model (uncertainty-aware loss + SAM optimizer). The evaluation metric is average DSC (\%), NSD (\%), and MSE.}
\label{tab:Results_ablation_study_Aorta}
\vspace{-2mm}
\centerline{
    % \normalsize
    \renewcommand{\arraystretch}{1.3} % Increase row height
    \scalebox{0.8}{
    \begin{tabular}{c|c|c|c}
        \hline
        Method                                 & Avg. DSC \% $\uparrow$ & Avg. NSD \% $\uparrow$ & Avg. MSE \% $\downarrow$ \\ \hline
        CE loss                                & 73.761                 & 92.141                 & 0.0960                   \\ \hline
        \makecell{Uncertainty-aware loss\\(CE, Dice, Focal)} & 75.053                 & 91.747                 & 0.0966                   \\ \hline
        \textbf{Ours}         & \textbf{77.084}        & \textbf{93.847}        & \textbf{0.0906}          \\ \hline
    \end{tabular}
    }
}
\vspace{-1mm}
\end{table}
%-------------------------------------
%-------------------------------------
\begin{table}[t]
\caption{Ablation study of ImageCAS dataset~\cite{zeng2023imagecas} on various configurations with U-Mamba backbone: (1) using only the Cross-Entropy (CE) loss, (2) using the uncertainty-aware Loss without the SAM optimizer, and (3) the proposed UU-Mamba model (uncertainty-aware loss + SAM optimizer). The evaluation metric is average DSC (\%), NSD (\%), and MSE.}
\label{tab:Results_ablation_study_ImageCAS}
\vspace{-2mm}
\centerline{
    \renewcommand{\arraystretch}{1.3} % Increase row height
    \scalebox{0.8}{
    \begin{tabular}{c|c|c|c}
        \hline
        Method                                   & Avg. DSC \% $\uparrow$ & Avg. NSD \% $\uparrow$ & Avg. MSE \% $\downarrow$ \\ \hline
        CE loss                                  & 79.496                 & 87.490                 & 0.0006784                \\ \hline
        \makecell{Uncertainty-aware loss\\(CE, Dice, Focal)} & 81.146                 & 88.045                 & 0.0006105                \\ \hline
        \textbf{Ours}           & \textbf{81.9983}       & \textbf{88.771}        & \textbf{0.0005903}       \\ \hline
    \end{tabular}
    }
}
\vspace{-1mm}
\end{table}
%-------------------------------------

%%%%%%%%%%%%%%%%%%%%%%%%%%%%%%%%%%%%%%%%%%%%%%%%%%%%%%%%%%%%%%%%%%%%%%%%%%%%
\subsection{Experimental Results}
\noindent We conduct a comparison of UU-Mamba with five of the state-of-the-art segmentation models—TransUNet~\cite{Chen2021TransUNet}, Swin-Unet~\cite{Cao2021SwinUnet}, nnUNet~\cite{Isensee2021nnUNet}, nnFormer~\cite{Zhou2021nnFormer}, and U-Mamba~\cite{ma2024umamba}—on the ACDC dataset~\cite{bernard2018deep}. Transformer-based networks are TransUNet and Swin-Unet, while nnUNet and nnFormer employ CNN-based architectures. U-Mamba is a hybrid architecture that combines components from both Transformer-based and CNN-based networks.

Using the Dice Similarity Coefficient (DSC) as the evaluation metric, we conducted a quantitative evaluation of UU-Mamba against these five 3D heart segmentation models on the ACDC dataset~\cite{bernard2018deep}. The segmentation results for the compared algorithms are depicted in Figure~\ref{fig:ACDC_result} on a few images from the ACDC dataset. Table~\ref{tab:Results_comparison_diff_models} provides the average DSC scores in all regions, as well as the DSC scores for each model in three cardiac regions: the right ventricle (RV), myocardium (Myo), and left ventricle (LV). Scores demonstrating the greatest performance are indicated in italics.

Our UU-Mamba model surpasses all other methods, attaining the highest overall performance with an average DSC of 92.79\%. The DSC scores for each region are as follows: 92.41\% for RV, 90.90\% for Myo, and 95.04\% for LV for each region. The flexibility and efficacy of our model are demonstrated by its exceptional ability to accurately segment the right ventricle and myocardium. Despite a slight decrease in DSC for the left ventricle compared to other models, this is counterbalanced by the highest overall average DSC and the superior scores in other regions, as illustrated in Table~\ref{tab:Results_comparison_diff_models}.

In contrast, TransUNet achieves an average DSC of 89.71\%, with a relatively lower score for Myo. The average DSC of Swin-Unet is 90.00\%, with the maximum DSC for LV and lower performance for RV. The nnUNet model achieves an average DSC of 91.61\%, with significant enhancements in Myo segmentation. Providing robust performance in all regions, particularly the myocardium, the nnFormer model obtains an average DSC of 92.06\%. The average DSC of U-Mamba of 92.22\% indicates substantial improvements in the RV and Myo regions.

The superior performance of our UU-Mamba model in comparison to the other existing models is emphasized by this quantitative evaluation. Our method exhibits the potential to improve the accuracy of 3D heart segmentation in medical imaging, as it has the highest average DSC and notably strong segmentation in the right ventricle and myocardium. The effectiveness of our approach is underscored by the enhancements it achieves over models such as U-Mamba, nnFormer, and nnUNet. This approach employs sophisticated loss functions and optimization techniques to capitalize on global and local features.

%-------------------------------------
\begin{table*}[t]
\caption{
Ablation study to compare DSC and NSD values for various anatomical zones and arteries using three methods: (1) only the Cross-Entropy (CE) loss, (2) the uncertainty-aware Loss without the SAM optimizer, and (3) the proposed UU-Mamba model. The best values for each region are highlighted in bold. The table demonstrates the effectiveness of the UU-Mamba model in achieving higher segmentation accuracy across most regions.
}
\label{tab:Results_ablation_study_Aorta_detail}
\vspace{-2mm}
\centerline{
    \scalebox{0.9}{
    \begin{tabular}{l|c|c|c|c|c|c}
        \hline
                                    &  \multicolumn{3}{c|}{DSC (\%)}                                               & \multicolumn{3}{c}{NSD (\%)}                                                \\ \hline
        Labels                      & CE loss           & \makecell{Uncertainty-\\aware loss} & UUMamba            & CE loss           & \makecell{Uncertainty-\\aware loss} & \textbf{Ours}           \\ \hline
        Zone 0                      & \textbf{89.317}   & 87.803                              & 87.604             & \textbf{76.749}   & 70.280                              & 72.291            \\ \hline
        Innominate                  & 75.151            & 77.830                              & \textbf{78.126}    & 81.251            & 84.259                              & \textbf{84.768}   \\ \hline
        Zone 1                      & \textbf{67.936}   & 65.432                              & 65.491             & \textbf{86.647}   & 84.379                              & 85.001            \\ \hline
        Left Common Carotid         & 78.001            & 77.547                              & \textbf{78.689}    & 92.852            & \textbf{93.539}                     & 93.453            \\ \hline
        Zone 2                      & 75.043            & \textbf{75.514}                     & 73.749             & \textbf{91.645}   & 91.627                              & 90.585            \\ \hline
        Left Subclavian Artery      & 83.260            & 83.578                              & \textbf{84.725}    & \textbf{99.124}   & 98.742                              & 98.612            \\ \hline
        Zone 3                      & \textbf{74.306}   & 73.513                              & 73.157             & \textbf{94.146}   & 92.939                              & 93.038            \\ \hline
        Zone 4                      & 79.590            & 84.233                              & \textbf{84.701}    & 89.008            & 91.987                              & \textbf{93.001}   \\ \hline
        Zone 5                      & 88.860            & 89.725                              & \textbf{89.780}    & 95.571            & 96.879                              & \textbf{97.271}   \\ \hline
        Zone 6                      & 71.166            & 71.119                              & \textbf{73.552}    & 97.692            & 94.694                              & \textbf{98.381}   \\ \hline
        Celiac Artery               & 67.016            & 66.311                              & \textbf{68.691}    & 97.270            & 96.138                              & \textbf{99.158}   \\ \hline
        Zone 7                      & 68.116            & 71.036                              & \textbf{74.067}    & 99.404            & 99.060                              & \textbf{99.622}   \\ \hline
        SMA                         & 69.002            & 70.117                              & \textbf{71.200}    & \textbf{88.940}   & 87.300                              & 88.905            \\ \hline
        Zone 8                      & 77.029            & \textbf{79.027}                     & 78.831             & \textbf{100.000}  & 99.576                              & \textbf{100.000}  \\ \hline
        Right Renal Artery          & \textbf{74.873}   & 72.898                              & 73.501             & \textbf{98.032}   & 96.949                              & 97.015            \\ \hline
        Left Renal Artery           & 67.907            & 70.187                              & \textbf{71.890}    & 96.650            & \textbf{98.323}                     & 97.640            \\ \hline
        Zone 9                      & 90.805            & 90.118                              & \textbf{91.008}    & \textbf{99.877}   & 99.636                              & 99.637            \\ \hline
        Right Common Iliac Artery   & 78.244            & 79.078                              & \textbf{86.141}    & 95.438            & 96.317                              & \textbf{98.783}   \\ \hline
        Left Common Iliac Artery    & 75.914            & 79.623                              & \textbf{86.457}    & 97.638            & 96.370                              & \textbf{99.864}   \\ \hline
        Right Internal Iliac Artery & 59.568            & 65.540                              & \textbf{66.409}    & 85.795            & 82.728                              & \textbf{88.424}   \\ \hline
        Left Internal Iliac Artery  & 67.034            & 64.154                              & \textbf{67.791}    & 96.124            & 90.865                              & \textbf{99.069}   \\ \hline
        Right External Iliac Artery & 59.216            & 61.344                              & \textbf{72.469}    & 77.545            & 77.270                              & \textbf{87.139}   \\ \hline
        Left External Iliac Artery  & 59.148            & 70.506                              & \textbf{74.895}    & 81.853            & 90.306                              & \textbf{96.825}   \\ \hline
    \end{tabular}
    }
}
\vspace{-1mm}
\end{table*}
%-------------------------------------
%-------------------------------------
\begin{figure}[t]
    \centerline{\includegraphics[width=\linewidth]{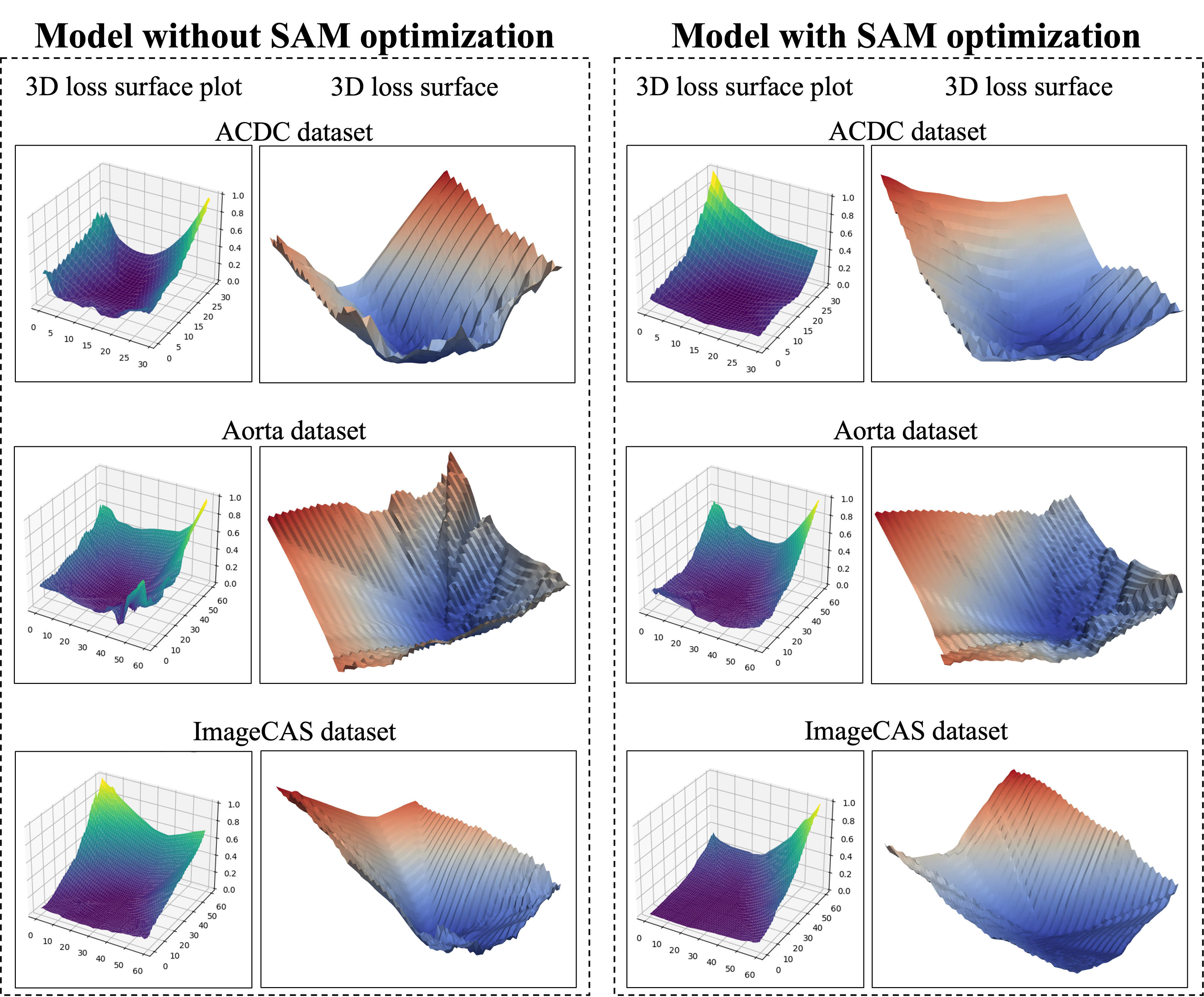}}
    % \centerline{\includegraphics[width=0.5\textwidth]{figure/SAM_alldataset2.png}}  
    \caption{Comparison of loss landscapes for models with and without SAM (Sharpness-Aware Minimization) optimization across three datasets: ACDC~\cite{bernard2018deep}, ImageCAS~\cite{zeng2023imagecas}, and Aorta~\cite{imran2024cis, krebs2024volumetric}. Each image represents the 3D loss landscape of a model trained on one of the datasets, illustrating the effect of SAM optimization on the model's ability to find flatter minima. The models without SAM exhibit sharper, more erratic loss contours, indicating less stable convergence, whereas the models with SAM show smoother, reflecting enhanced generalization capabilities. This visualization highlights the impact of SAM optimization in improving model robustness and training stability across diverse data conditions.}
    \vspace{-3mm}
    \label{fig:SAM_results}
\end{figure}
%-------------------------------------

%%%%%%%%%%%%%%%%%%%%%%%%%%%%%%%%%%%%%%%%%%%%%%%%%%%%%%%%%%%%%%%%%%%%%%%%%%%%
\subsection{Ablation Study}
\noindent We perform an ablation study to investigate the impact of integrating Sharpness-Aware Minimization (SAM) optimization~\cite{foret2021sharpness} into our UU-Mamba model, alongside traditional and uncertainty-aware loss functions. This study spans three datasets: ACDC~\cite{bernard2018deep}, Aorta~\cite{imran2024cis, krebs2024volumetric}, and ImageCAS~\cite{zeng2023imagecas}, and employs 3D loss surface visualizations~\cite{Hao2018} using ParaView~\cite{ahrens2005ParaView, ayachit2015, ayachit2021} to demonstrate the smoother loss landscapes indicative of model robustness and improved generalization.

\subsubsection{Impact of Traditional and Uncertainty-Aware Loss Functions}
\noindent To evaluate the impact of different loss functions on model performance, we first assess the baseline performance using the standard Cross-Entropy (CE) loss. This traditional approach achieves satisfactory results but has limitations in handling complex segmentation challenges. To address these, we introduce an uncertainty-aware loss function, which combines Dice loss, CE loss, and Focal loss. This combination provides a more balanced approach by emphasizing confident predictions and reducing the negative impact of uncertain areas in the segmentation process.

%%% ACDC Dataset %%%
In the ACDC dataset~\cite{bernard2018deep}, the model trained with only CE loss achieves a Dice Similarity Coefficient (DSC) of 92.263\%. When incorporating the uncertainty-aware loss, the DSC improves to 92.602\%. This enhancement reflects the efficacy of the uncertainty-aware approach in improving segmentation resilience by carefully managing prediction uncertainty and refining the model's focus on areas of high confidence, as also shown in Table \ref{tab:Results_ablation_study_ACDC}.

%%% Aorta Dataset %%%
For the Aorta dataset~\cite{imran2024cis, krebs2024volumetric}, training with only CE loss yields an average Dice Similarity Coefficient (DSC) of 73.761\%. Incorporating the uncertainty-aware loss results in notable improvements across various anatomical zones and arteries, raising the average DSC to 75.053\% and average NSD to 91.747\%, also reducing the Mean Squared Error (MSE). These improvements, as shown in Table \ref{tab:Results_ablation_study_Aorta}, highlight the uncertainty-aware loss’s ability to handle complex segmentation tasks by balancing confident predictions with uncertain areas.

%%% ImageCAS Dataset %%%
For the ImageCAS dataset~\cite{zeng2023imagecas}, the model trained with only CE loss achieves an average Dice Similarity Coefficient (DSC) of 79.496\%. By incorporating the uncertainty-aware loss, the average DSC improves to 81.146\% and NSD improves to 88.045\%, demonstrating the advantage of this comprehensive loss function in enhancing segmentation accuracy, particularly in challenging regions, as shown in Table \ref{tab:Results_ablation_study_ImageCAS}.

\subsubsection{Enhancements with Sharpness-Aware Minimization optimization}
\noindent Building on the incorporation of uncertainty-aware loss, we further integrate Sharpness-Aware Minimization (SAM) optimization~\cite{foret2021sharpness} to explore its additional benefits. SAM is designed to steer the training process toward flatter minima, which are associated with improved generalization in neural network models. Figure~\ref{fig:SAM_results} illustrates a comparison of the loss landscapes between models trained with and without SAM optimization.

%%% ACDC Dataset %%%
In Table \ref{tab:Results_ablation_study_ACDC}, incorporating SAM with the uncertainty-aware loss increases the ACDC dataset's DSC to 92.787\%, the highest among the tested methods, thus validating SAM's role in enhancing segmentation precision and generalization.
In the Figure~\ref{fig:SAM_results}, the loss landscape on the ACDC dataset is shown for models with and without SAM optimization. The 3D loss surface plot of the model without SAM optimization exhibits a broader range of loss values, characterized by sharper and more erratic loss contours. In contrast, the model utilizing SAM optimization displays a flatter and smoother loss landscape, indicating improved stability and generalization.

%%% Aorta Dataset %%%
As shown in Table \ref{tab:Results_ablation_study_Aorta}, incorporating SAM into the model training process increases the average DSC to 77.084\%, along with significant improvements in both DSC and NSD metrics. SAM optimization leads to smoother and more stable loss surfaces, as demonstrated in Figure~\ref{fig:SAM_results}. Without SAM, the 3D loss surface exhibits sharper and more rugged contours, particularly along the edges. In contrast, the model with SAM displays a flatter, more stable loss landscape, indicating improved robustness and generalization.

Notably, the UU-Mamba model achieves the highest DSC scores in 17 out of 24 anatomical regions and the highest NSD values in 13 out of 24 regions, as detailed in Table~\ref{tab:Results_ablation_study_Aorta_detail}. These results underscore the superior generalizability and accuracy of the UU-Mamba model. The consistent top performance across most regions highlights the significant benefits of combining uncertainty-aware loss with SAM optimization to enhance segmentation outcomes.

%%% ImageCAS Dataset %%%
As detailed in Table \ref{tab:Results_ablation_study_ImageCAS}, SAM optimization pushes the ImageCAS dataset performance metrics to the highest levels observed in this study. In Figure~\ref{fig:SAM_results}, the 3D loss surface of the model without SAM shows greater variability, especially in the bottom right region. By contrast, with SAM optimization, the loss surface becomes much smoother, indicating improved model consistency, robustness, and generalization across diverse cardiovascular imaging scenarios.

Incorporating Sharpness-Aware Minimization (SAM) optimization significantly improves performance across various cardiovascular imaging datasets. SAM promotes flatter minima in the training process, leading to better generalization and segmentation precision. In the ACDC dataset, SAM boosts the DSC to 92.787\%, the highest among tested methods. For the Aorta and ImageCAS datasets, SAM enhances both DSC and NSD metrics, leading to smoother and more stable loss landscapes, reflecting improved model consistency and performance across diverse datasets.

%%%%%%%%%%%%%%%%%%%%%%%%%%%%%%%%%%%%%%%%%%%%%%%%%%%%%%%%%%%%%%%%%%%%%%%%%%%%
\subsection{Robustness Analysis}
\noindent We perform experiments to evaluate each method on the Mean Squared Error (MSE) of the DSC scores to assess their robustness quantitatively. The MSE is calculated as shown in Eq.~\eqref{eq:MSE}. Results are shown in the Tables \ref{tab:Results_ablation_study_ACDC}, \ref{tab:Results_ablation_study_Aorta}, and \ref{tab:Results_ablation_study_ImageCAS}. These results show that the uncertainty-aware loss reduces the MSE compared to the standard CE loss, reflecting its ability to better address the variability and uncertainty in the data. The inclusion of SAM optimization significantly decreases the MSE, achieving the lowest error value. This reduction in MSE highlights SAM's role in minimizing errors and producing more accurate segmentation maps.

Figures \ref{fig:ACDC_result}, \ref{fig:Aorta_result}, and \ref{fig:ImageCAS_result} show the MSE between the output segmentation and the ground truth for each method, providing a visual comparison of the segmentation quality. These visualizations complement the quantitative results by illustrating the error distribution and highlighting areas where the SAM optimization and uncertainty-aware loss contribute to more accurate and consistent segmentation outcomes.

%%%%%%%%%%%%%%%%%%%%%%%%%%%%%%%%%%%%%%
%%%%%%%%%%%%%%%%%%%%%%%%%%%%%%%%%%%%%%
\section{Conclusion}
\noindent We present a novel model, UU-Mamba, specifically developed for the purpose of segmenting cardiovascular MRI and CT data. This model combines the U-Mamba architecture with an uncertainty-aware loss function and the SAM optimizer, resulting in a substantial enhancement of biological picture segmentation. It achieves improved generalization and boundary accuracy. The uncertainty-aware loss function integrates region-based, distribution-based, and pixel-based losses to enhance segmentation performance by effectively managing jobs and prioritizing confident predictions. Simultaneously, the SAM optimizer directs the model towards flat minima in the loss landscape, improving its ability to withstand challenges and decreasing the likelihood of overfitting, ultimately resulting in more accurate segmentation.

Aside from doing our main tests on the ACDC dataset~\cite{bernard2018deep}, we also assessed the performance of UU-Mamba on two other datasets: ImageCAS~\cite{zeng2023imagecas} and Aorta~\cite{imran2024cis, krebs2024volumetric}. The model scored the greatest average DSC, NSD, and MSE on the ImageCAS dataset, demonstrating a considerable improvement compared to the baseline models. UU-Mamba demonstrated superior performance compared to other models on the Aorta dataset, earning the greatest average DSC in 17 out of 24 anatomical regions and the highest average NSD in 13 out of 24 anatomical regions. The data illustrate that the model is highly effective in different anatomical regions and segmentation tasks, highlighting its versatility and strength in numerous medical imaging situations.
The comparative analysis conducted on five prominent models establishes the superiority of UU-Mamba. It achieves a DSC of 92.787\% on the ACDC dataset, demonstrating high accuracy and robustness in segmenting various datasets, such as ImageCAS and Aorta.

Future work will involve examining supplementary data augmentation strategies, exploring other ways for modeling uncertainty, and validating the model on bigger and more varied datasets. Our main objective is to improve and expand the UU-Mamba technique in order to enhance automated medical imaging.

%\subsection{Acknowledgments}
%\noindent Xin Wang is supported by the University at Albany Start-up Grant. 
%The authors appreciate the computational resource provided by the University at Albany -- SUNY. 

\small
\bibliographystyle{IEEEtran}
\bibliography{reference}

\end{document}